\documentclass{article}

\usepackage{graphicx}
\usepackage{subfigure}
\usepackage{algorithm}
\usepackage{algorithmic}
\usepackage{authblk}
\usepackage{amsmath}
\usepackage{amssymb}
\usepackage{amsthm}
\usepackage{color} 
\usepackage{hyperref}
\usepackage{chngpage}

\DeclareMathOperator*{\argmax}{arg\,max}
\DeclareMathOperator*{\argmin}{arg\,min}

\newcommand{\innerprod}[2]{{\left\langle {#1},{#2} \right\rangle}}
\newcommand{\norm}[1]{{\left\lVert {#1} \right\rVert}}
\newcommand{\functional}[2]{{\mathcal{#1}} [#2]}
\newcommand{\set}[1]{\mathcal{#1}}
\newcommand{\etal}{\textit{et al.}}
\newcommand{\eg}{\textit{e.g.}}
\newcommand{\ie}{\textit{i.e.}}
\newcommand{\SpeedBoost}{\textsc{SpeedBoost}}
\newcommand{\StructuredSpeedBoost}{\textsc{StructuredSpeedBoost}}
\newcommand{\HIM}{\textsc{HIM}}
\newcommand{\SIM}{\textsc{SIM}}
\newcommand\addtag{\refstepcounter{equation}\tag{\theequation}}
\definecolor{ctaRoad}{RGB}{210, 105, 30}
\definecolor{ctaFacade}{RGB}{205, 92, 92}
\definecolor{ctaTrunk}{RGB}{160, 82, 45}
\definecolor{ctaVehicle}{RGB}{223, 0, 255}
\definecolor{ctaVeg}{RGB}{34, 139, 87}
\definecolor{ctaPole}{RGB}{0, 0, 205}

\definecolor{sbdSky}{RGB}{128, 128, 128}
\definecolor{sbdTree}{RGB}{128, 128, 0}
\definecolor{sbdRoad}{RGB}{128, 64, 128}
\definecolor{sbdGrass}{RGB}{0, 128, 0}
\definecolor{sbdWater}{RGB}{0, 0, 128}
\definecolor{sbdBuilding}{RGB}{128, 0, 0}
\definecolor{sbdMountain}{RGB}{128, 64, 0}
\definecolor{sbdObject}{RGB}{192, 128, 0}
\newcommand{\SBDColors}
{\begin{tabular}{cccccccc}
\colorbox{sbdSky}{\textcolor{sbdSky}{|}} Sky &
\colorbox{sbdTree}{\textcolor{sbdTree}{|}} Tree &
\colorbox{sbdRoad}{\textcolor{sbdRoad}{|}} Road &
\colorbox{sbdGrass}{\textcolor{sbdGrass}{|}} Grass &
\colorbox{sbdWater}{\textcolor{sbdWater}{|}} Water &
\colorbox{sbdBuilding}{\textcolor{sbdBuilding}{|}} Building &
\colorbox{sbdMountain}{\textcolor{sbdMountain}{|}} Mountain &
\colorbox{sbdObject}{\textcolor{sbdObject}{|}} Object
\end{tabular}
}

\definecolor{camvidBuilding}{RGB}{128, 0, 0}
\definecolor{camvidTree}{RGB}{128, 128, 0}
\definecolor{camvidSky}{RGB}{128, 128, 128}
\definecolor{camvidCar}{RGB}{64, 0, 128}
\definecolor{camvidSignSymbol}{RGB}{192, 128, 128}
\definecolor{camvidRoad}{RGB}{128, 64, 128}
\definecolor{camvidPedestrian}{RGB}{64, 64, 0}
\definecolor{camvidFence}{RGB}{64, 64, 128}
\definecolor{camvidColumnPole}{RGB}{192, 192, 128}
\definecolor{camvidSidewalk}{RGB}{0, 0, 192}
\definecolor{camvidBicyclist}{RGB}{0, 128, 192}
\newcommand{\CamVidColors}
{
\begin{tabular}{c}
\colorbox{camvidBuilding}{\textcolor{camvidBuilding}{|}} Building ~
\colorbox{camvidTree}{\textcolor{camvidTree}{|}} Tree ~
\colorbox{camvidSky}{\textcolor{camvidSky}{|}} Sky ~
\colorbox{camvidCar}{\textcolor{camvidCar}{|}} Car ~
\colorbox{camvidSignSymbol}{\textcolor{camvidSignSymbol}{|}} Sign ~
\colorbox{camvidRoad}{\textcolor{camvidRoad}{|}} Road ~
\colorbox{camvidPedestrian}{\textcolor{camvidPedestrian}{|}} Person ~
\colorbox{camvidFence}{\textcolor{camvidFence}{|}} Fence ~
\colorbox{camvidColumnPole}{\textcolor{camvidColumnPole}{|}} Pole ~
\colorbox{camvidSidewalk}{\textcolor{camvidSidewalk}{|}} Sidewalk ~
\colorbox{camvidBicyclist}{\textcolor{camvidBicyclist}{|}} Bicyclist
\end{tabular}
}

\definecolor{TXTColor}{RGB}{0, 255, 0}
\definecolor{LBPColor}{RGB}{255, 0, 255}
\definecolor{ISIFTColor}{RGB}{255, 0, 0}
\definecolor{CSIFTColor}{RGB}{0, 0, 255}
\newcommand{\FeatureColors}
{
\scriptsize
\begin{tabular}{cccc}
\colorbox{TXTColor}{\textcolor{TXTColor}{|}} \texttt{TXT} &
\colorbox{ISIFTColor}{\textcolor{ISIFTColor}{|}} \texttt{I-SIFT} &
\colorbox{CSIFTColor}{\textcolor{CSIFTColor}{|}} \texttt{C-SIFT} &
\colorbox{LBPColor}{\textcolor{LBPColor}{|}} \texttt{LBP}
\end{tabular}
}

\title{SpeedMachines: Anytime Structured Prediction}
\author{Alexander Grubb\thanks{agrubb@cmu.edu}}
\author{Daniel Munoz\thanks{dmunoz@cs.cmu.edu}}
\author{J. Andrew Bagnell\thanks{dbagnell@cs.cmu.edu}}
\author{Martial Hebert\thanks{hebert@cs.cmu.edu}}
\affil{School of Computer Science, Carnegie Mellon University, Pittsburgh, PA 15213 USA}
\date{June 20, 2013}

\begin{document}

\maketitle

\begin{abstract}
Structured prediction plays a central role in machine
learning applications from computational biology to
computer vision. These models require significantly
more computation than unstructured models, and, in many
applications, algorithms may need to make predictions
within a computational budget or in an anytime fashion.
In this work we propose an anytime technique for
learning structured prediction that, at training time, incorporates both
structural elements and feature computation trade-offs that affect test-time inference.
We apply our technique to the challenging
problem of scene understanding in computer vision and demonstrate efficient and
anytime predictions that gradually improve towards state-of-the-art classification
performance as the allotted time increases.
\end{abstract}

\section{Introduction}
In real-world applications, we are often forced to trade-off between
accurate predictions and the computation time needed to make them.  In
many problems, structured prediction algorithms are necessary to
obtain accurate predictions; however, this class of techniques is
typically more computationally demanding over simpler locally
independent predictions.  Furthermore, under limited computational
resources we may forced to make a prediction after a limited and
unknown amount of time.  Therefore, we require an approach that is
efficient while also capable of returning a sensible prediction when
requested at any time.  Furthermore, as the inference procedure is
given more time, we should expect the predictive performance to also
increase.

The contribution of this work is an algorithm for making \emph{anytime} structured predictions.  As detailed
in the following sections, our approach accounts for both inference
and feature computation times while automatically trading-off cost for
accuracy, and \textit{vice versa}, in order to maximize predictive performance
with respect to computation time. Although our approach is applicable
towards a variety of different structured prediction problems, we
analyze its efficacy on the challenging problem of scene
understanding in computer vision.  Our experiments
demonstrate that we can learn an efficient, anytime predictor whose classification
performance improves towards state-of-the-art while automatically
selecting what features to compute and where to compute them with respect to time.

\subsection{Related Work}
A canonical approach for incorporating computation time during
learning is a cascade of feed-forward modules, where each module
becomes more sophisticated but also more computationally expensive the
further it is down the cascade \cite{viola-ijcv-04}.  One drawback of
the cascade approach is that the procedure is trained for a specific
sequence length and is not suited for interruption.  For example,
stopping predictions after one module in a long cascade will generally
perform much worse than a predictor that was trained in isolation.
Recent works \cite{gao-nips-11, grubb-aistats-12, xu-icml-12, karayev-nips-12}
have investigated techniques for
learning locally independent predictors which balance feature computation time and
inference time during learning; however, they are not immediately
amenable to the structured prediction scenario.

In the structured setting, Jiang \etal\ \cite{jiang-nips-12} proposed a technique for
reinforcement learning that incorporates a user specified speed/accuracy trade-off distribution,
and Weiss and Taskar \cite{weiss-aistats-10}
proposed a cascaded analog for structured prediction where the
solution space is is iteratively refined/pruned over time.  In
contrast, we are focused on learning a structured predictor with
interruptible, anytime properties which is also trained to balance
both the structural and feature computation times during the inference procedure.
Recent work in computer vision and
robotics \cite{sturgess-bmvc-12, nijs-iros-12} has similarly
investigated techniques for making approximate inference in graphical
models more efficient via a cascaded procedure that iteratively prunes
subregions in the scene to analyze.  We similarly incorporate such
structure selection in our approach; however, we also account for
feature computation time and avoid the early hard commitments required
with a cascaded approach.  This allows for early predictions to be as
accurate as possible, as they do not have to be made conservatively as
in the cascade approach.

\section{Background}
\subsection{Structured Prediction}
In the structured prediction setting, we are given inputs $x \in
\set{X}$ and associated structured outputs $y \in \set{Y}$.  The goal
is to learn a function $f : \set{X} \rightarrow \set{Y}$ that
minimizes some risk $\functional{R}{f}$, typically evaluated pointwise
over the inputs:
\begin{equation}
\label{eq:pointwise-loss}
\functional{R}{f} = \mathbb{E}_{\set{X}}[l(f(x))].
\end{equation}

We will further assume that each input and output pair has some
underlying structure, such as the graph structure of graphical models,
that can be utilized to predict portions of the output locally.  Let
$j$ index these structural elements.  We then assume that a final
structured output $y$ can be represented as a variable length vector
$(y_1, \ldots, y_J)$, where each element $y_j$ lies in some vector
space $y_j \in \set{Y}'$.  For example, these outputs could be the
probability distribution over class labels for each pixel in an image,
or distributions of part-of-speech labels for each word in a sentence.
Similarly we can compute some features $x_j$ representing the portion
of the input which corresponds to a given output, such as features
computed over a neighborhood around a pixel in an input image.

As another example, consider labeling tasks such as part-of-speech
tagging.  In this domain, we are given a set of input sentences
$\set{X}$, and for each word $j$ in a given sentence, we want to
output a vector $\hat{y}_j \in \mathbb{R}^K$ containing the scores
with respect to each of the $K$ possible part-of-speech labels for
that word.  This sequence of vectors for each word is the complete
structured prediction $\hat{y}$.  An example loss function for this
domain would be the multiclass log-loss, averaged over words, with
respect to the ground truth parts-of-speech.

Along with the encoding of the problem, we also assume that the
structure can be used to reduce the scope of the prediction problem,
as in graphical models.  One common approach to generating predictions
on these structures is to use a policy-based or iterative decoding
approach, instead of probabilistic inference over a graphical model
\cite{cohen-ijcai-05, daume-mlj-09, tu-pami-10, socher-icml-11,
  ross-cvpr-11}.  In order to model the contextual relationships among
the outputs, these iterative approaches commonly perform a sequence of
predictions, where each update relies on previous predictions made
across the structure of the problem.

Let $N(j)$ represent the locally connected elements of $j$, such as
the locally connected factors of a node $j$ in a typical graphical
model.  For a given node $j$, the predictions over the neighboring
nodes $\hat{y}_{N(j)}$ can then be used to update the prediction for
that node.  For example, in the character recognition task, the
predictions for neighboring characters can influence the prediction
for a given character, and be used to update and improve the accuracy
of that prediction.

In the iterative decoding approach a predictor $\phi$ is iteratively
used to update different elements $\hat{y}_j$ of the final structured
output:
\begin{equation}
\hat{y}_j = \phi(x_j, \hat{y}_{N(j)}).
\end{equation}

A complete policy then consists of a strategy for selecting which elements
of the structured output to update, coupled with the predictor for
updating the given outputs.  Typical approaches include randomly selecting
elements to update, iterating over the structure in a fixed ordering,
or simultaneously updating all predictions at all iterations.  As shown
by Ross \etal\ \cite{ross-cvpr-11}, this iterative decoding approach can
is equivalent to message passing approaches used to solve graphical models,
where each update encodes a single set of messages passed to one node
in the graphical model.

\subsection{Anytime Prediction}
In traditional boosting, the goal is to learn a function
\begin{equation}
f(x) = \sum_t \alpha_t h_t(x),
\end{equation}
which is additively built from a set of weaker predictors $h \in \set{H}$
that minimizes some risk functional
$\argmin_{f \in \set{F}} \functional{R}{f}$.
Minimizing this functional can be viewed as performing gradient
descent in function space \cite{mason-lmc-99, friedman-as-00}.
Assuming the loss function is of the form given in (Eq.~\ref{eq:pointwise-loss}), the
functional gradient, $\nabla = \nabla_f \functional{R}{f}$, is a function of the form
\begin{equation}
\nabla(x) = \frac{\partial l(f(x))}{\partial f(x)}.
\end{equation}
The weak predictor $h$ that best minimizes the projection error of the functional gradient
is selected at each iteration:
\begin{align}
h_t &= \argmax_{h \in \mathcal{H}} \innerprod{\nabla \functional{R}{f_{t-1}}}{h}, \\
\alpha_t &= \argmin_{\alpha \in \mathbb{R}} \functional{R}{f_{t-1} + \alpha h_t}.
\end{align}
Minimizing this projection error can be equivalently performed through the least
squares minimization
\begin{equation}
\label{eq:vanilla-lsq}
h_t = \argmin_{h \in \set{H}} \mathbb{E}_{\set{X}}\left[ \|\nabla(x) - h(x)\|^2 \right].
\end{equation}

Extending this framework, Grubb and Bagnell \cite{grubb-aistats-12} introduce
an anytime prediction method that modifies the standard boosting criterion to
automatically trade-off the loss of a weak predictor, $h$, with its cost $c(h) \in \mathbb{R}^+$.
They do this by using a cost-greedy selection criterion
that selects the weak predictor which gets the best improvement in
loss per unit cost,
\begin{equation}
\label{eq:speedboost}
h_t, \alpha_t = \argmax_{h \in \mathcal{H}, \alpha \in \mathbb{R}}
    \frac{\mathcal{R}\left[f_{t-1}\right] - \mathcal{R}\left[f_{t-1} + \alpha h \right]}{c(h)}.
\end{equation}
Hence, for some fixed number of iterations $T$,
the total cost of the learned function is
$c(f_T) = \sum_{t=1}^T c(h_t)$.
The learned predictor can easily adjust to any new budget of costs by evaluating the
sequence until the budget is exhausted.
Grubb and Bagnell prove that this \SpeedBoost\ algorithm updates
the resulting predictions at an increasing sequence of budgets
that is competitive with any other sequence which uses the same
weak predictors for a wide range of budgets \cite{grubb-aistats-12}.

\section{Anytime Structured Prediction}
\subsection{Weak Structured Predictors}
We adapt the \SpeedBoost\ framework to the structured prediction
setting by learning an additive structured predictor.  To accomplish this,
we will adapt the policy-based iterative decoding approach to use
an additive policy instead of one which replaces previous predictions.

In the iterative decoding described previously, recall that we have two components,
one for selecting which elements to update, and another for updating the predictions
of the given elements.  Let $S^t$ be the set of components
selected for updating at iteration $t$.  For current predictions $y^t$ we can re-write the policy
for computing the predictions at the next iteration of the iterative decoding procedure
as:
\begin{align}
\hat{y}^{t+1}_j &=
  \begin{cases}
   \phi(x_j, \hat{y}^t_{N(j)}) & \text{if } j \in \set{S}^t \\
   \hat{y}^t_j                 & \text{otherwise}
  \end{cases}.
\end{align}

The additive version of this policy instead uses weak predictors $h$, each of which maps both
the input data and previous structured output to a more refined
structured output, $h :\set{X} \times \set{Y} \rightarrow \set{Y}$:
\begin{equation}
\hat{y}^{t+1} = \hat{y}^t + h(x, \hat{y}^t).
\end{equation}

We can build a weak predictor $h$ which performs the same actions as
the previous replacement policy by considering weak predictors with two parts:
a function $h_{\textrm{S}}$ which selects which structural elements to
update, and a predictor $h_{\textrm{P}}$ which runs on the selected
elements and updates the respective pieces of the structured output.

The selection function $h_{\textrm{S}}$ takes in an input $x$ and
previous prediction $\hat{y}$ and outputs a set of structural nodes
$\set{S} = \{j_1, j_2, \ldots\}$ to update.  For each structural
element selected by $h_{\textrm{S}}$, the predictor $h_{\textrm{P}}$
takes the place of $\phi$ in the previous policy, taking $(x_j,
\hat{y}_{N(j)})$ and computing an update for the prediction
$\hat{y}_j$.

Returning to the part-of-speech tagging example, possible selection
functions would select different chunks of the sentence, either
individual words or multi-word phrases using some selection criteria.
Given the set of selected elements, a prediction function would take
each selected word or phrase and update the predicted distribution over the
part-of-speech labels using the features for that word or phrase.

Using these elements we can write the weak predictor $h$, which
produces a structured output $(h(\cdot)_1, \ldots,
h(\cdot)_J)$, as
\begin{align}
h(x, \hat{y}^t)_j &=
  \begin{cases}
   h_{\textrm{P}}(x_j, \hat{y}^t_{N(j)}) & \text{if } j \in h_{\textrm{S}}(x, \hat{y}) \\
   0                                     & \text{otherwise}
  \end{cases},
\end{align}
or alternatively we can write this using an indicator function:
\begin{equation}
\label{eq:weak-learner}
h(x,\hat{y}^t)_j = \mathbf{1}(j \in h_{\textrm{S}}(x, \hat{y}^t)) h_{\textrm{P}}(x_j, \hat{y}^t_{N(j)}).
\end{equation}

The adapted cost model for this weak predictor is then simply the
sum of the cost of evaluating both the selection function and the
prediction function,
\begin{equation}
\label{eq:cost}
c(h) = c(h_{\textrm{S}}) + c(h_{\textrm{P}}).
\end{equation}

\subsection{Selecting Weak Predictors}
In order to use the greedy improvement-per-unit-time selection
strategy used by \SpeedBoost\ in (Eq.~\ref{eq:speedboost}), we need to
be able to complete the projection operation over the $\set{H}$.  We
assume that we are given a fixed set of possible selection functions,
$\set{H}_{\textrm{S}}$, and a set of $L$ learning algorithms, $\{
\mathcal{A}_l \}_{l=1}^L$, where $\mathcal{A} : \set{D} \rightarrow
\set{H}_{\textrm{P}}$ generates a predictor given a training set $D$.
In practice, these algorithms are generated by varying the
complexities of the algorithm, \eg, depths in a decision tree
predictor.

Given a fixed selection function $h_{\textrm{S}}$ and
current predictions $\hat{y}$, we can build a dataset
appropriate for training weak predictors $h_{\textrm{P}}$ as follows.
In order to minimize the projection error in (Eq.~\ref{eq:vanilla-lsq})
for a predictor $h$ of the form in (Eq.~\ref{eq:weak-learner}),
it can be shown that this reduces to finding the prediction function $h_{\textrm{P}}$ that minimizes
\begin{equation}
\label{eq:project-lsq}
\argmin_{h_{\textrm{P}} \in \set{H_{\textrm{P}}}}
  \mathbb{E}_{\set{X}}\left[ \sum_{j \in h_{\textrm{S}}(x, \hat{y})}
    \norm{\nabla(x)_j - h_{\textrm{P}}(x_j, \hat{y}_{N(j)})}^2 \right],
\end{equation}
where
\begin{equation}
\nabla(x)_j = \frac{\partial l(f(x))}{\partial f(x)_j},
\end{equation}
the gradient of the loss with respect to the partial structured prediction
$\hat{y}_j$.

This optimization problem is equivalent to minimizing weighted least squares
error over the dataset
\begin{align}
D &= \bigcup_{x} \bigcup_{j \in h_{\textrm{S}}(x, \hat{y})}
       \{ \left( \psi_j, \nabla(x)_j \right) \},\\
  &= \textrm{gradient}(f, h_{\textrm{S}}), \addtag \label{eq:gradient-dataset}
\end{align}
where $\psi_j = \psi(x_j, \hat{y}_{N(j)})$ is the feature descriptor
for the given structural node, and $\nabla(x)_j$ is its target.  In
order to model contextual information, $\psi$ is drawn from both the
raw features $x_j$ for the given element and the previous locally
neighboring predictions $\hat{y}_{N(j)}$.

Algorithm~\ref{alg:structured-speedboost} summarizes the
\StructuredSpeedBoost\ algorithm for anytime structured prediction.
It enumerates the candidate selection functions, $h_{\textrm{S}}$,
creates the training dataset defined by (Eq.~\ref{eq:gradient-dataset}), and then
generates a candidate prediction function $h_{\textrm{P}}$ using each
weak learning algorithm.  For all the pairs of candidates,
it uses the \SpeedBoost\ criteria to select the most
cost efficient pair, and then repeats.

\begin{algorithm}[tb]
  \caption{\StructuredSpeedBoost}
  \label{alg:structured-speedboost}
  \begin{algorithmic}
    \STATE {\textbf{Given:}} objective $\mathcal{R}$,
    cost function $c$,
    set of selection functions $\set{H}_{\textrm{S}}$,
    set of $L$ learning algorithms $\{\mathcal{A}_l\}_{l=1}^L$,
    number of iterations $T$,
    initial function $f_0$.
     \FOR{$t = 1,\ldots,T$}
    \STATE $\set{H}^* = \emptyset$
        \FOR{$h_{\textrm{S}} \in \set{H}_{\textrm{S}}$}

    \STATE Create dataset
    $D = \textrm{gradient}(f_{t-1}, h_{\textrm{S}})$ (Eq.~\ref{eq:gradient-dataset})
    \FOR{$\mathcal{A} \in \{\mathcal{A}_1, \ldots , \mathcal{A}_L \}$}

        \STATE Train $h_{\textrm{P}} = \mathcal{A}(D)$

        \STATE Define $h$ from $h_{\textrm{S}}$ and $h_{\textrm{P}}$ (Eq.~\ref{eq:weak-learner})

        \STATE $\set{H}^* = \set{H}^* \cup \{h\}$
        \ENDFOR
    \ENDFOR
    \STATE $h_t, \alpha_t =
      \argmax_{h \in \mathcal{H}^*, \alpha \in \mathbb{R}}
      \frac{\mathcal{R}\left[f_{t-1}\right] - \mathcal{R}\left[f_{t-1} + \alpha h \right]} {c(h)}$
    \STATE $f_t = f_{t-1} + \alpha_t h_t$
  \ENDFOR
  \end{algorithmic}
\end{algorithm}

\subsection{Handling Limited Training Data}
In the presence of limited training data, training the prediction
function $h_{\textrm{P}}$ using previous predictions $\hat{y}$ can
lead to overfitting.  In practice, this can be alleviated by adapting
the concept of stacking \cite{wolpert-nn-92}, which has demonstrated
to be useful in other structured prediction work \cite{cohen-ijcai-05,
  munoz-eccv-10}.  Conceptually, the idea is that we do not want to
use the same $f$ we are currently learning to generate $\hat{y}$ for
use in the next boosting iteration.  Concretely, we can instead split
our entire training set $\Omega$ into two disjoint subsets, $\Omega =
A \cup B$, $A \cap B = \emptyset$. At training time, we learn three
separate structured predictors $f, f^A, f^B$ over datasets $\Omega, A,
B$, respectively.  Let $h_{\textrm{P}}, h_{\textrm{P}}^A,
h_{\textrm{P}}^B$ be the prediction functions for the structured
predictors $f, f^A, f^B$, respectively.  When training
$h_{\textrm{P}}$ over $x \in \Omega$, the predictions $\hat{y}$ are
generated from held-out predictions: for $x \in A$, $\hat{y} =
f^B(x)$, and for $x \in B$, $\hat{y} = f^A(x)$.  Now we need to train
$f^A$ and $f^B$ in the same hold-out manner: when training
$h_{\textrm{P}}^A$ over $x \in A$, $\hat{y} = f^B(x)$, and when
training $h_{\textrm{P}}^B$ over $x \in B$, $\hat{y} = f^A(x)$.  This
interleaving process is solely done at training time to induce
less-overfit predictions $\hat{y}$ when training structured predictor
$f$.  Since $f$ is trained over all the training data, we use solely
its prediction at test-time and discard $f^A$ and $f^B$. In practice,
we follow this procedure using 10 folds instead of just two.


\begin{figure*}[!t]
\centering
\begin{tabular}{cc}
\includegraphics[width=0.2\linewidth]{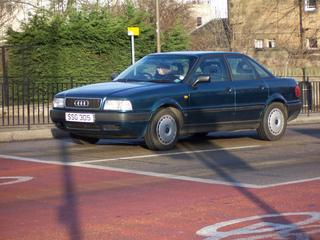} &
\includegraphics[width=0.5\linewidth]{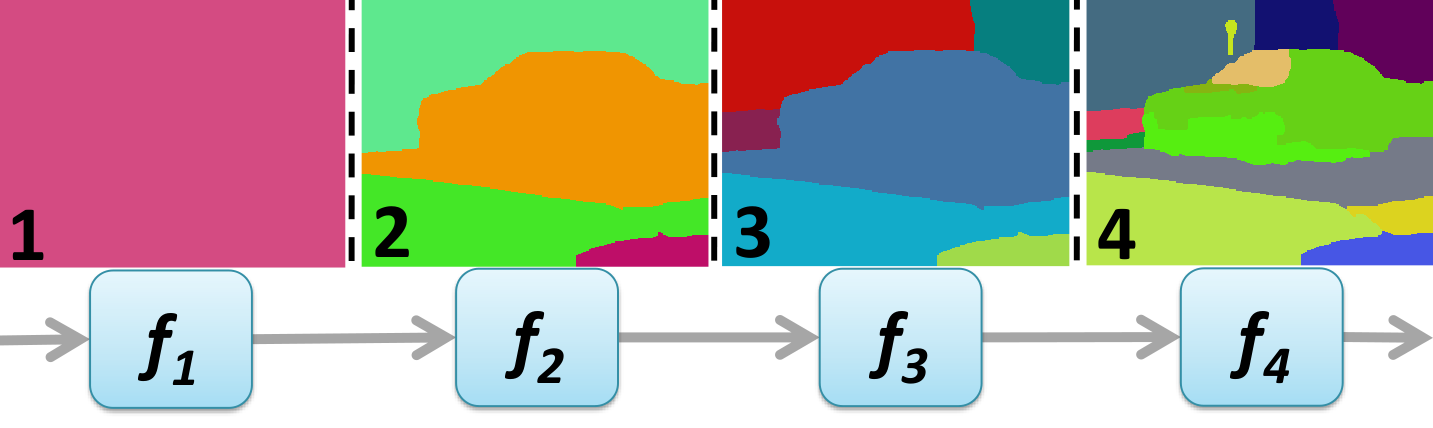} \\
(a) & (b)
\end{tabular}
\caption{
Hierarchical Inference Machines \cite{munoz-eccv-10}.
(a) Input image.
(b) The image is segmented multiple times; predictions are made and passed between levels.
Images courtesy of the authors' ECCV 2010 presentation.
}
\label{fig:him}
\end{figure*}

\section{Anytime Scene Understanding}
\subsection{Background}
In addition to part-of-speech tagging in natural language processing,
scene understanding in computer vision is another important and
challenging structured prediction problem.  The \textit{de facto}
approach to this problem is with random field based models
\cite{kumar-ijcv-06, gould-ijcv-08, ladicky-eccv-10}, where the random
variables in the graph represent the object category for a
region/patch in the image.  While random fields provide a clean
interface between modeling and inference, recent works
\cite{tu-pami-10, munoz-eccv-10, socher-icml-11, fabaret-icml-12} have
demonstrated alternative approaches that achieve equivalent or
improved performances with the additional benefit of a simple,
efficient, and modular inference procedure.

Inspired by the hierarchical representation used in the
state-of-the-art scene understanding technique from Munoz
\etal\ \cite{munoz-eccv-10}, we apply \StructuredSpeedBoost\ to
the scene understanding problem by reasoning over differently sized
regions in the scene.  In the following, we briefly review the
hierarchical inference machine (\HIM) approach from
\cite{munoz-eccv-10} and then describe how we can perform an anytime
prediction whose structure is similar in spirit.

\subsection{Hierarchical Inference Machines}
\HIM\ parses the scene using a hierarchy of segmentations, as
illustrated in Fig.~\ref{fig:him}.  By incorporating multiple
different segmentations, this representation addresses the problem of
scale ambiguity in images.  Instead of performing (approximate)
inference on a large random field defined over the regions, inference
is broken down into a sequence of predictions.  As illustrated in
Fig.~\ref{fig:him}, a predictor $f$ is associated with each level in
the hierarchy that predicts the probability distribution of
classes/objects contained within each region.  These predictions are
then used by the subsequent predictor in the next level (in addition
to features derived from the image statistics) to make refined
predictions on the finer regions; and the process iterates.  By
passing class distributions between predictors, contextual information
is modeled even though the segmentation at any particular level may be
incorrect.  We note that while Fig.~\ref{fig:him} illustrates a
top-down sequence over the hierarchy, in practice, the authors iterate
up and down the hierarchy which we also do in our comparison
experiments.

\subsection{Speedy Inference Machines}
While \HIM\ decomposes the structured prediction problem into an efficient sequence of predictions,
it is not readily suited for an anytime prediction.
\textbf{First}, the final predictions are generated when the procedure
terminates at the leaf nodes in the hierarchy.  Hence, interrupting the procedure before then
would result in final predictions over coarse regions that
may severely undersegment the scene.
\textbf{Second}, the amount of computation time at each step of the procedure is invariant to the current
performance.  Because the structure of the sequence is predefined, the inference procedure
will predict multiple times on a region as it traverses over the hierarchy, even though there may
be no room for improvement.
\textbf{Third}, the input to each predictor in the sequence is a fixed feature descriptor for the region.
Because these input descriptors must be precomputed for all regions in the hierarchy
before the inference process begins, there is a fixed initial computational cost.
In the following, we describe how \StructuredSpeedBoost\ addresses these three problems three problems
for anytime scene understanding.

\subsubsection{Interruptible Prediction}
In order to address the first issue, we learn an additive predictor
$f$ which predicts a per-pixel classification for the entire image at once.
In contrast to \HIM\ whose \emph{multiple} predictors' losses are measured over regions,
we train a \emph{single} predictor whose loss is measured over pixels.
Concretely, given per-pixel ground truth distributions $p_j \in \mathbb{R}^K$,
we wish to optimize per-pixel, cross-entropy risk for all pixels in the image
\begin{equation}
\mathcal{R}[f] = \mathbb{E}_{\set{X}}\left[ -\sum_j \sum_k p_{jk} \log q(f(x))_{jk} \right],
\end{equation}
where
\begin{equation}
q(y)_{jk} = \frac{\exp(y_{jk})} {\sum_{k'} \exp(y_{jk'})},
\end{equation}
\ie, the probability of the $k$'th class for the $j$'th pixel.
Using (Eq.~\ref{eq:weak-learner}), the probability distribution associated with each pixel
is then dependent on
1) the pixels to update, selected by $h_{\textrm{S}}$,
and 2) the value of the predictor $h_{\textrm{P}}$ evaluated on those respective pixels.
The definition of these functions are defined in the following subsections.

\begin{table*}[!t]
\centering
\tabcolsep 3.0pt
\scalebox{0.8}{
\begin{tabular}{@{}l|llllllll|ll@{}}
    & \rotatebox{90}{Sky}
    & \rotatebox{90}{Tree}
    & \rotatebox{90}{Road}
    & \rotatebox{90}{Grass}
    & \rotatebox{90}{Water}
    & \rotatebox{90}{Bldg.}
    & \rotatebox{90}{Mtn.}
    & \rotatebox{90}{Object}
    & \rotatebox{90}{\textit{Class}}
    & \rotatebox{90}{\textit{Pixel}} \\
    \hline
    \SIM\           & 92.4 & 73.3 & 90.6 & 83.0 & 62.5 & 76.9 & 10.5 & 63.8 & 69.1 & 78.8 \\
    \HIM\ \cite{munoz-eccv-10}	& 92.9  & 77.5 & 89.9 & 83.6 & 70.9 & 83.2 & 17.2 & 69.3 & 73.1 & 82.1 \\
    \hline
    \cite{fabaret-icml-12}  & 95.7 & 78.7 & 88.1 & 89.7 & 68.7 & 79.9 & 44.6 & 62.3 & 76.0 & 81.4 \\
    \cite{socher-icml-11}   & - & - & - & - & - & - & - & - & - & 78.1
\end{tabular}
}
\vspace{.5cm}\\
\scalebox{0.8}{
\begin{tabular}{@{}l|lllllllllll|ll@{}}
    & \rotatebox{90}{Bldg.}
    & \rotatebox{90}{Tree}
    & \rotatebox{90}{Sky}
    & \rotatebox{90}{Car}
    & \rotatebox{90}{Sign}
    & \rotatebox{90}{Road}
    & \rotatebox{90}{Ped.}
    & \rotatebox{90}{Fence}
    & \rotatebox{90}{Pole}
    & \rotatebox{90}{Sdwlk.}
    & \rotatebox{90}{Bike}
    & \rotatebox{90}{\textit{Class}}
    & \rotatebox{90}{\textit{Pixel}} \\
    \hline
    \SIM\				& 76.8 & 79.2 & 94.1 & 71.8 & 30.2 & 95.2 & 34.6 & 25.7 & 14.0 & 66.3 & 15.0 & 54.8 & 81.5 \\
    \HIM\ \cite{munoz-eccv-10}		& 83.3 & 82.2 & 95.9 & 75.2 & 42.2 & 96.0 & 38.6 & 21.5 & 13.6 & 72.1 & 33.3 & 59.4 & 84.9 \\
    \hline
    \cite{nijs-iros-12}         & 59   & 75   & 93   & 84   & 45   & 90   & 53   & 27   & 0    & 55   & 21   & 54.7 & 75.0 \\
    \cite{ladicky-eccv-10}$^\dagger$	& 81.5 & 76.6 & 96.2 & 78.7 & 40.2 & 93.9 & 43.0 & 47.6 & 14.3 & 81.5 & 33.9 & 62.5 & 83.8
\end{tabular}
}
\caption{Recalls on the Stanford Background Dataset (top) and CamVid (bottom)
where \textit{Class} is the average per-class recall and \textit{Pixel} is the per-pixel accuracy.
$^\dagger$Uses additional training data not leveraged by other techniques.}
\label{tab:leaderboard}
\end{table*}

\subsubsection{Structure Selection and Prediction}
In order to account for scale ambiguity and structure in the scene, we
can similarly integrate multiple regions into our predictor.  By using
a hierarchical segmentation of the scene that produces many
segments/regions, we can consider each resulting region or segment of pixels $S$
in the hierarchy as one possible set of outputs to update.  Intuitively, there is no
need to update regions of the image where the predictions are correct
at the current inference step.  Hence, we want to update the portion
of the scene where the predictions are uncertain, \ie, have high
entropy $H$.  To achieve this, we use a selector function that selects
regions that have high average per-pixel entropies in the current
predictions,
\begin{equation}
h_{\textrm{S}}(x, \hat{y}) = \left\{ S \ \middle|\  \frac{1}{|S|} \sum_{j \in S} H(q(\hat{y})_j) > \theta\right\},
\end{equation}
for some fixed threshold $\theta$. In practice, the set of predictors $\set{H}_{\textrm{S}}$
used at training time is created from a diverse set of thresholds $\theta$.

Additionally, we assume that the features $\psi_j$ used for each pixel
in a given selected region are drawn from the entire region, so that
if a given scale is selected features corresponding to that scale are
used to update the selected pixels.  For a given segment $S$, call
this feature vector $\psi_S$.

Given the above selector function, we use (Eq. \ref{eq:project-lsq}) to find the
next best predictor function, as in Algorithm~\ref{alg:structured-speedboost}, optimizing
\begin{equation}
\label{eq:grad-project-basic}
 h_{\textrm{P}}^* =
\underset{h_{\textrm{P}}}{\arg \min}
\sum_{S \in h_{\textrm{S}}(x, \hat{y})} \sum_{j \in S} \norm{\nabla(x)_j - h_{\textrm{P}}(\psi_S)}^2.
\end{equation}

Because
all pixels in a given region use the same feature vector, this reduces to the
weighted least squares problem:
\begin{equation}
\label{eq:grad-project}
 h_{\textrm{P}}^* =
\underset{h_{\textrm{P}}}{\arg \min}
\sum_{S \in h_{\textrm{S}}(x, \hat{y})} |S| \norm{\nabla_S - h_{\textrm{P}}(\psi_S)}^2.
\end{equation}
where $\nabla_S = \mathbb{E}_{j \in S}[\nabla(x)_j] = \mathbb{E}_{j \in S}[p_j - q(\hat{y})_j]$.
In words, we find a vector-valued regressor $h_{\textrm{P}}$ with minimal weighted least squares error
between the difference in ground truth and predicted per-pixel distributions,
averaged over each selected region/segment, and weighted by the size of the selected region.
This is an intuitive update that places large weight to updating large regions.

\subsubsection{Dynamic Feature Computation}
In the scene understanding problem, a significant computational
cost during inference is often feature descriptor computation.  To this end, we
utilize the \SpeedBoost\ cost model (Eq.~\ref{eq:cost}) to automatically
select the most computationally efficient features.

The features used in this application, drawn from previous work
\cite{gould-ijcv-08, ladicky-thesis-11} and detailed in the following section,
are computed as follows.  First, a set of base feature descriptors are
computed from the input image data.
In many applications it is useful to quantize these base feature descriptors and pool
them together to form a set of derived features \cite{coates-aistats-11}.
We follow the soft vector quantization
approach in \cite{coates-aistats-11} to form a quantized code vector
by computing distances to multiple cluster centers in a dictionary.

\begin{table*}[t!]
\centering
\begin{adjustwidth}{-1in}{-1in}
\begin{tabular}{c|c|cc|cc|cc|cc}
\texttt{FH} & \texttt{SHAPE} &
\texttt{TXT} (B) & \texttt{TXT} (D) &
\texttt{LBP} (B) & \texttt{LBP} (D) &
\texttt{I-SIFT} (B) & \texttt{I-SIFT} (D) &
\texttt{C-SIFT} (B) & \texttt{C-SIFT} (D) \\
\hline
 167 & 2 & 29 & 66 & 64  & 265 & 33 & 165 & 93 & 443 
\end{tabular}
\end{adjustwidth}
\caption{Average timings (ms) for computing all features for an image in the SBD.
(B) is the time to compute the \emph{base} per-pixel feature responses,
and (D) is the time to compute the \emph{derived} averaged-pooled region codes to all cluster centers.}
\label{tab:timings}
\end{table*}

This computation incurs a fixed cost for 1) each group of features with a
common base feature, and 2) an additional, smaller fixed cost for
each actual feature used.  In order to account for these costs, we use
an additive model similar to Xu \etal\ \cite{xu-icml-12}.
Formally, let $\phi \in \Phi$ be the set of features and $\gamma \in
\Gamma$ be the set of feature groups, and $c_\phi$ and $c_\gamma$ be the cost
for computing derived feature $\phi$ and the base feature for group
$\gamma$, respectively.  Let $\Phi(f)$ be the set of features used by
predictor $f$ and $\Gamma(f)$ the set of its used groups.  Given a current predictor
$f_{t-1}$, its group and derived feature costs are then
just the costs of any new group and derived features and have not previously been computed:
\begin{align}
c_{\Gamma}(h_{\textrm{P}}) &= \sum_{\gamma \in \Gamma(h_{\textrm{P}}) \setminus \Gamma(f_{t-1})} c_{\gamma},\\
c_{\Phi}(h_{\textrm{P}}) &= \sum_{\phi \in \Phi(h_{\textrm{P}}) \setminus \Phi(f_{t-1})} c_{\phi}.
\end{align}
The total cost model in (Eq. \ref{eq:cost}) can then be derived using the sum
of the feature costs and group costs as
\begin{align}
\label{eq:costz}
c(h) &= c(h_{\textrm{s}}) + c(h_{\textrm{P}}) \\
     &= \epsilon_{\textrm{S}} + \epsilon_{\textrm{P}} + c_{\Gamma}(h_{\textrm{P}}) + c_{\Phi}(h_{\textrm{P}}) \addtag,
\end{align}
where $\epsilon_{\textrm{S}}$ and $\epsilon_{\textrm{P}}$ are
small fixed costs for evaluating a selection and prediction function, respectively.

In order to generate $h_{\textrm{P}}$ with a variety of costs, we
use a modified regression tree that penalizes each split based on its
potential cost, as in \cite{xu-icml-12}.  This approach augments
the least-squares regression tree impurity function with a cost regularizer:
\begin{equation}
\mathbb{E}_{D} \left[ w_D \norm{ y_D - h_{\textrm{P}}(x_D) }^2 \right] +
  \lambda \left( c_{\Gamma}(h_{\textrm{P}}) + c_{\Phi}(h_{\textrm{P}}) \right),
\end{equation}
where $\lambda$ regularizes the cost.
In addition to (Eq.~\ref{eq:costz}),
training regression trees with different values of $\lambda$,
enables \StructuredSpeedBoost\ to automatically select the most
cost-efficient predictor.

\section{Experimental Analysis}
\subsection{Setup}
We evaluate performance metrics between \SIM\ and \HIM\
on the
1) Stanford Background Dataset (SBD) \cite{gould-iccv-09},
which contains 8 classes, and
2) Cambridge Video Dataset (CamVid) \cite{brostow-eccv-08},
which contains 11 classes; we follow the same training/testing
evaluation procedures as originally described in the respective papers.
As shown in Table \ref{tab:leaderboard}, we note that \HIM\ achieves state-of-the-art
performance and these datasets and analyze the computational tradeoffs when compared with \SIM.
Since both methods operate over a region hierarchy of the scene, we
use the same segmentations, features, and regression trees (weak predictors) for a fair comparison.

\subsubsection{Segmentations} We construct a 7-level segmentation hierarchy by recursively
executing the graph-based segmentation algorithm (\texttt{FH}) \cite{fh-ijcv-04} with parameters
\begin{align*}
\sigma &= 0.25, c = 10^2 \times [1, 2, 5, 10, 50, 200, 500], \\
k &= [30, 50, 50, 100, 100, 200, 300].
\end{align*}
These values were qualitatively chosen to generate regions at different resolutions.

\subsubsection{Features}
\label{sec:experiments-features}
A region's feature descriptor is composed of 5 feature groups ($\Gamma$):
1) region boundary shape/geometry/location (\texttt{SHAPE}) \cite{gould-ijcv-08},
2) texture (\texttt{TXT}),
3) local binary patterns (\texttt{LBP}),
4) SIFT over intensity (\texttt{I-SIFT}),
5) SIFT separately over colors R, G, and B (\texttt{C-SIFT}).
The last 4 are derived from per-pixel descriptors for which we use the
publicly available implementation from \cite{ladicky-thesis-11}.

\begin{figure*}[t!]
\centering
\includegraphics[width=0.8\textwidth]{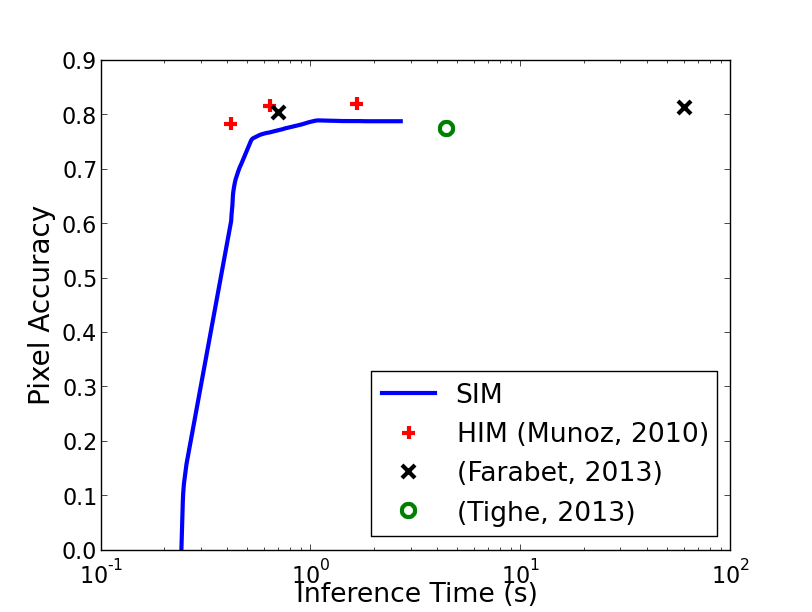}\\
\includegraphics[width=0.8\textwidth]{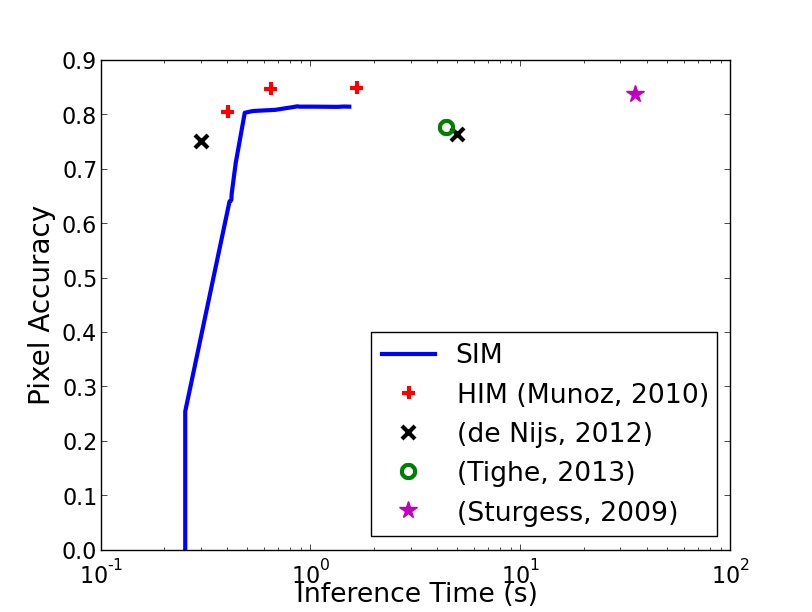}
\caption{Average pixel classification accuracy for SBD (top) and CamVid (bottom) datasets as a function of inference time.}
\label{fig:anytime-f1s}
\end{figure*}

Computations for segmentation and features are shown in Table \ref{tab:timings};
all times were computed on an Intel i7-2960XM processor.
The \texttt{SHAPE} descriptor is computed solely from the segmentation boundaries
and is efficient to compute.  The remaining 4 feature group computations are
broken down into the per-pixel descriptor (base) and the average-pooled vector quantized
codes (derived), where each of the 4 groups are quantized separately with a
dictionary size of 150 elements/centers using k-means.
For a given pixel descriptor, $\upsilon$, its code assignment
to cluster center, $\mu_i$, is derived from its squared $L_2$ distance
$d_i(\upsilon) = \norm{\upsilon - \mu_i}_2^2$.
Using the soft code assignment from \cite{coates-aistats-11}, the code is defined as $\max(0, z_i(\upsilon))$, where
\begin{align}
z_i(\upsilon) &= \mathbb{E}_j[d_j(\upsilon)] - d_i(\upsilon)\\
 &= \mathbb{E}_j[\|\mu_j\|^2] - 2 \innerprod{\mathbb{E}_j[\mu_j]}{\upsilon} - (\|\mu_i\|^2 - 2 \innerprod{\mu_i}{\upsilon}).
\end{align}
Note that the expectations are indepndent from the query descriptor $v$, hence
the $i$'th code can be computed independently and enables selective computation for the region.
The resulting quantized pixel codes are then averaged within each region.
Thus, the costs to use these derived features are dependent if the pixel
descriptor has already been computed or not. For example, when the
weak learner \emph{first} uses codes from the \texttt{I-SIFT} group,
the cost incurred is the time
to compute the \texttt{I-SIFT} pixel descriptor \emph{plus} the time to compute
distances to each specified center.

\begin{figure}[t!]
\centering
\begin{tabular}{c}
\subfigure{\includegraphics[width=1.0\textwidth]{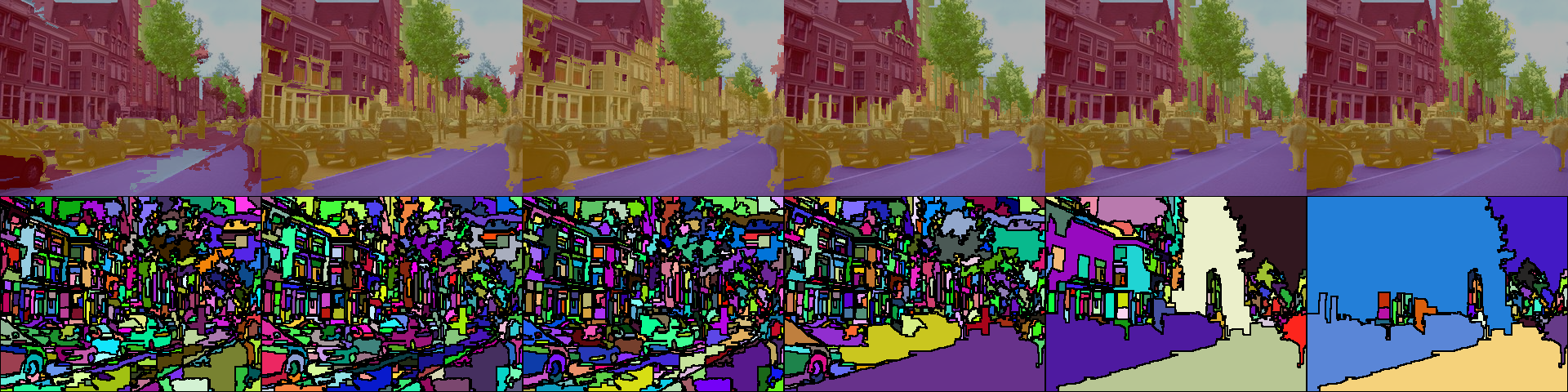}}
\end{tabular}
\begin{adjustwidth}{-1in}{-1in}
\centering
{\scriptsize \SBDColors }
\end{adjustwidth}
\begin{tabular}{c}
\subfigure{\includegraphics[width=1.0\textwidth]{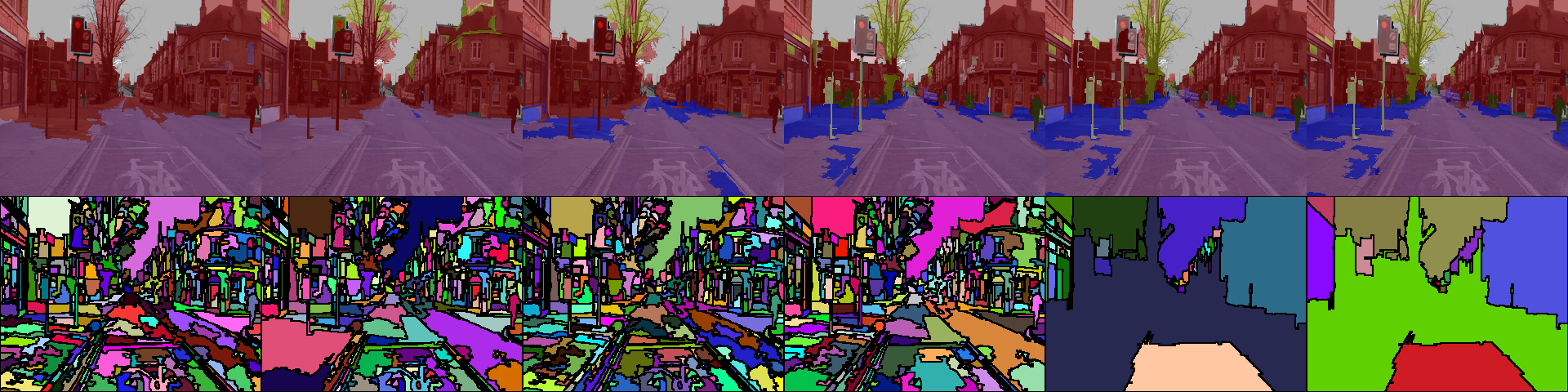}}
\end{tabular}
\begin{adjustwidth}{-1in}{-1in}
\centering
{\scriptsize \CamVidColors }
\end{adjustwidth}
\caption{Sequence of images displaying the inferred labels and selected
regions at iterations $t = \{1, 5, 15, 50, 100, 225\}$ of the \SIM\ algorithm
for a sample image from the Stanford Background (top) and CamVid (bottom) datasets.
The corresponding inference times for these iterations are
$\{0.42s, 0.44s, 0.47s, 0.79s, 1.07s, 1.63s\}$ (top) and
$\{0.41s, 0.42s, 0.44s, 0.52s, 0.85s, 1.42s\}$ (bottom).}
\label{fig:pretty-images}
\end{figure}

\begin{figure}[t]
\centering
\includegraphics[width=0.8\textwidth]{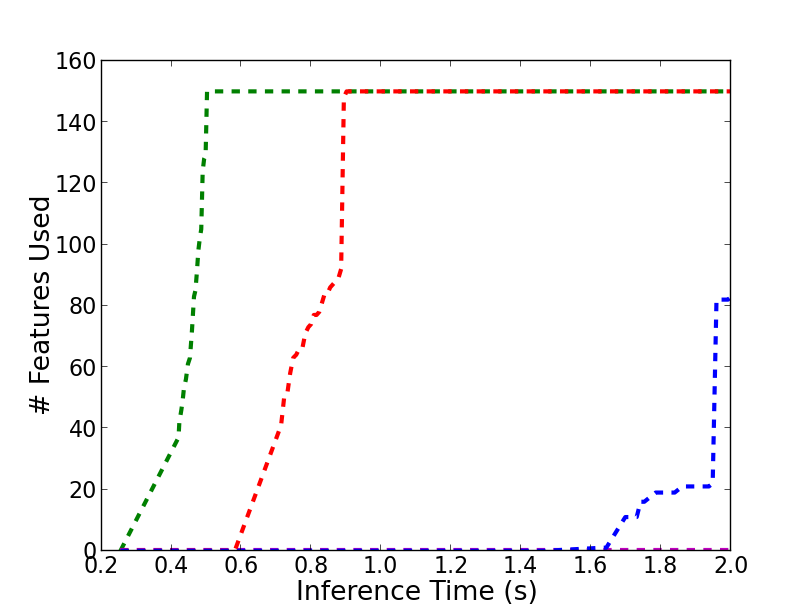}\\
\FeatureColors
\caption{The number of cluster centers selected within each feature group by
\SIM\ as a function of inference time.}
\label{fig:sbd-features-used}
\end{figure}

\subsection{Analysis}
In Fig.~\ref{fig:sbd-features-used} we show which cluster centers, from
each of the four groups, are being selected by \SIM\ as the inference
time increases.  We note that efficient \texttt{SHAPE} descriptor is
chosen on the first iteration, followed by the next cheapest
descriptors \texttt{TXT} and \texttt{I-SIFT}.  Although
\texttt{LBP} is cheaper than \texttt{C-SIFT}, the algorithm
ignored \texttt{LBP} because it did not improve prediction
wrt cost.

In Fig.~\ref{fig:anytime-f1s}, we compare the classification performance of
\SIM\ and several other algorithms with respect to inference time.
We consider \HIM\ as well as two variants which use a limited set
of the 4 feature groups (only \texttt{TXT} and \texttt{TXT} \& \texttt{I-SIFT});
these \SIM\ and \HIM\ models were executed on the same computer.
We also compare to the reported performances of other techniques and
stress that these timings are reported from different computing configurations.
The single anytime
predictor generated by our anytime structured prediction approach is
competitive with all of the specially trained, standalone models without
requiring any of the manual analysis necessary to create the different
fixed models.

\nocite{tighe-ijcv-13, sturgess-bmvc-09}


In Fig.~\ref{fig:pretty-images}, we show the progress of the \SIM\ algorithm
as it processes a scene from each of the datasets.  Over time, we see
the different structural nodes (regions) selected by the algorithm as well
as improving classification.


\section{Conclusion}
We proposed a technique for structured prediction with anytime properties.
Our approach is based under the boosting framework that automatically incorporates new
learners to our predictor that best increases performance with respect to efficiency in terms
of both feature and inference computation times. We demonstrated the efficacy of our approach
on the challenging task of scene understanding in computer vision and achieved state-of-the-art performance
classifications with improved efficiency over previous work.


\bibliographystyle{plain}
\bibliography{references}
\end{document}